# A Fault Prognostic System for the Turbine Guide Bearings of a Hydropower Plant Using Long-Short Term Memory (LSTM)

Yasir Saleem Afridi[1], Mian Ibad Ali Shah[2*], Adnan Khan[1], Atia Kareem[1], Laiq Hasan[1]
Department of Computer Systems Engineering,
University of Engineering & Technology Peshawar, Peshawar, Pakistan[1]
School of Computer Science,
University of Galway, Galway, Ireland[2]

## ABSTRACT

Hydroelectricity, being a renewable source of energy, globally fulfills the electricity demand. Hence, Hydropower Plants (HPPs) have always been in the limelight of research. The fast-paced technological advancement is enabling us to develop state-of-the-art power generation machines. This has not only resulted in improved turbine efficiency but has also increased the complexity of these systems. In lieu thereof, efficient Operation & Maintenance (O&M) of such intricate power generation systems has become a more challenging task. Therefore, there has been a shift from conventional reactive approaches to more intelligent predictive approaches in maintaining the HPPs. The research is therefore targeted to develop an artificially intelligent fault prognostics system for the turbine bearings of an HPP. The proposed method utilizes the Long Short-Term Memory (LSTM) algorithm in developing the model. Initially, the model is trained and tested with bearing vibration data from a test rig. Subsequently, it is further trained and tested with realistic bearing vibration data obtained from an HPP operating in Pakistan via the Supervisory Control and Data Acquisition (SCADA) system. The model demonstrates highly effective predictions of bearing vibration values, achieving a remarkably low RMSE.

## 1. INTRODUCTION

In world energy consumption, hydropower contributes an astonishing 11,300 TWh [1]. Keeping in view such large share, the optimization of hydropower generation and operations becomes notably consequential, encompassing economic and societal dimensions.

The advantages of optimizing hydropower operations are multifaceted and extend beyond the enhancement of energy production. They encompass augmented energy security, mitigation of equipment failures and downtimes, extension of remaining useful life (RUL) for plant and equipment, among others. Notably, machine downtimes resulting from faults trigger an expansion in the demand-supply gap, a scenario with grave consequences. Hence, the necessity for cutting-edge Operation and Maintenance (O&M) systems becomes imperative. Nevertheless, attaining the aspiration of optimal hydropower plant (HPP) operation is far from straightforward.

In the realm of continuous production systems, such as electricity generation facilities, the cost incurred due to production loss during plant unavailability amid faults is exceedingly substantial. Considering this, well-devised maintenance procedures play a pivotal role in sustaining seamless equipment operation throughout its economic lifespan. These maintenance methodologies fall broadly into three categories: corrective, preventive, and predictive maintenance. In the context of Hydropower Plants (HPPs), the prevalent maintenance practices predominantly encompass preventive and corrective approaches [2].

The corrective approach is reactive in nature, entailing maintenance actions solely subsequent to the occurrence of faults. This conventional strategy for fault rectification is widely embraced [9]. On the contrary, preventive maintenance involves scheduled periodic upkeep of equipment, undertaken to diminish the likelihood of its malfunctioning [9]. This proactive method is performed while the equipment is still operational, ensuring it doesn't encounter unforeseen breakdowns. Conversely, prognostics revolves around the anticipation of faults and failures, aiming to predict when a system or component will cease to execute its intended function [9]. Given its efficacy in curbing unwarranted downtimes, prognostic maintenance has emerged as a focal point of contemporary research efforts.

## 1.1 MACHINE PROGNOSTICS

Prognostics can be categorized into two primary domains: data-driven approaches and model-based/physics-based approaches [24, 28, 36]. In physics-based methodologies, the development of models integrates domain expertise along with measured data through mathematical equations, considering the underlying physical laws. However, these physics-based approaches come with certain drawbacks. Firstly, their performance heavily hinges on the accuracy and quality of domain knowledge [24]. In practical scenarios, obtaining high-caliber domain knowledge can be challenging due to complexities and noisy operational conditions, subsequently impacting the model's robustness. Secondly, many of these models struggle to perform effectively in real-time settings, thereby curtailing their adaptability [23].

Conversely, data-driven models discern patterns from historical data, enabling informed decisions based on sensor-derived information [27]. Moreover, the proliferation of advanced computing devices and sophisticated sensors heightens the appeal of these data-driven condition monitoring systems [29]. Therefore, this research integrates a data-driven framework that leverages the data acquired from sensors as input to the model. Sensor data consists of time series data that are sampled and presented sequentially [24]. The data is processed by an algorithm in two distinct phases: firstly, the model is trained using historical data, and subsequently, the model is tested using current data.

## 1.2 USE OF BEARING DATA

The data is collected through various sensors strategically placed on critical components of machinery. Among these components, bearings play a crucial role, particularly in heavy-load mechanical systems like turbines [7, 8, 14, 21]. Bearings are vital for guiding and supporting rotating machine shafts. Given the demanding operational conditions, even a minor fault in bearings can lead to catastrophic consequences for the machinery, resulting in substantial financial losses [2]. Consequently, there exists a paramount need to swiftly detect bearing faults and embrace a proactive maintenance approach [6, 20]. Considering the pivotal role of bearings in plant operations and the frequency of related faults, this research concentrates on designing a prognostic maintenance system. The objective is to forecast bearing faults by analyzing the vibration data gathered from embedded sensors. The proposed solution employs an Artificial Intelligence (AI) model built on the LSTM algorithm. The primary goal is to effectively anticipate and predict faults occurring in hydropower turbine bearings.

## 1.3 LITERATURE REVIEW

Researchers globally have been actively engaged in forecasting bearing failures using diverse AI techniques, including Machine Learning (ML) algorithms like Deep Neural Networks (DNNs). Numerous studies have applied these methods to prognostic maintenance in renewable energy systems. AI stands out in comparison to conventional data processing methods due to its proficiency in modeling the unpredictable and non-stationary characteristics of nonlinear data [23]. Therefore, comprehensive research has been conducted on the application of various ML algorithms, including Artificial Neural Networks (ANNs) and DNNs, for predictive maintenance purposes. This research encompasses algorithms like Feed Forward Neural Network, Recurrent Neural Network (RNN), and Convolutional Neural Network (CNN), and has been extensively explored [34]. CNNs have the capability to autonomously capture features and generate meaningful representations of time series data, thereby eliminating the need for manual feature engineering [4]. Capability of 1D CNN time-series forecasting has been explored in [4]. 1D-CNN and BiLSTM implementation tutorial have been presented in [38] to predict peak electricity demand and price. Recently, deep learning has captured the interest of researchers due to its capability to model intricate nonlinear traits, effectively extracting intrinsic structures and valuable features from raw data [18]. Literature underscores the remarkable potential of deep learning across diverse domains, including Natural Language Processing (NLP) [10], Computer Vision (CV) [11], and Fault Diagnosis [12]. Other than the application in CV and NLP, various types of transformers have been explored in [36, 37] which are being used to interpret time-series modeling. Similarly, DNNs have shown encouraging outcomes in the prediction of renewable energy system prognosis [23]. Chen et al. [13] developed a CNN-based deep learning algorithm for carrying out gearbox fault diagnosis.

As previously mentioned, bearings, being integral to rotating machinery, operate under strenuous conditions, leading to a relatively high occurrence of faults. Any malfunction in bearings can result in production losses, equipment damage, and potential safety risks [14]. However, timely detection of these bearing faults enhances machinery reliability and performance [3].

Numerous techniques have been explored in the literature to accurately detect bearing faults. In the realm of time series analysis, RNNs exhibit favorable performance. Nevertheless, conventional RNNs struggle with extended sequences due to issues like gradient vanishing and exploding [15]. This challenge is effectively mitigated by LSTM networks, which excel in managing long-term data dependencies [5, 22]. Jiujian Wang et al [25] developed a Bidirectional LSTM (BiLSTM) model for estimating the Remaining Useful Life (RUL) of bearings using the C-MAPSS turbofan engine dataset, by NASA. Cheng-Geng Huang et al [26] and Zhao R et al [27] also developed a novel prognostics framework based on BiLSTM networks for achieving more accurate estimation of RUL and prediction of engineered systems that are subject to complex operational condition. Likewise, Jianling Qu et al [3] developed a stacked LSTM model for identification and recognition of faults in rolling bearings.

Unlike the models specified earlier in the section, that either uses time-domain features for training and testing the model or carry their analysis in frequency domain, this research is focused on developing a fault prognostics model that uses raw vibration signals generated by the turbine guide bearings of a real hydro power project. This will not only help in reducing the need for extensive domain knowledge but will also help in better generalization of the model.

The rest of this paper is organized as follows. Section 2 provides a detailed description of the methodology including an introduction to the LSTM model and its comparison with other models, section 3 discusses the results generated in the research, whereas section 4 concludes the paper.

## 2. METHODOLOGY

The research is structured into two main phases. Firstly, the developed model undergoes training and testing using bearing vibration data generated by a test rig. This dataset is sourced from the Prognostic Data Repository of NASA. The dataset was made available to be used publicly by the Center of Intelligent Maintenance System (IMS) at the University of Cincinnati [32]. Secondly, the model is trained and tested with authentic vibration data from bearings collected through the SCADA system at the Neelum Jhelum Hydro Power Project (NJHPP), a 969MW facility operating in Pakistan. The results are assessed using the Root Mean Square Error (RMSE) as an evaluative metric. Figure 1 visually outlines the comprehensive

## 2.1. Model Selection (Why LSTM?):

RNNs were originally introduced to address time sequence learning challenges [2,15]. While conventional neural networks are structured as multilayer networks capable of mapping input data solely to target vectors, RNNs have the unique capability of retaining information from previous inputs throughout the sequence. Like many other neural networks, RNNs utilize the backpropagation algorithm for training. However, they encounter a challenge during backpropagation known as vanishing gradients [19]. This issue arises as gradients shrink when propagated backward through time, dwindling to a point where their impact on learning becomes minimal. Consequently, traditional RNNs face limitations in research methodology. The subsequent section will delve into the rationale behind selecting, the architecture, and the operational principles of the LSTM algorithm. This will be followed by a detailed account of data acquisition, data preprocessing, and the model's training and testing process.

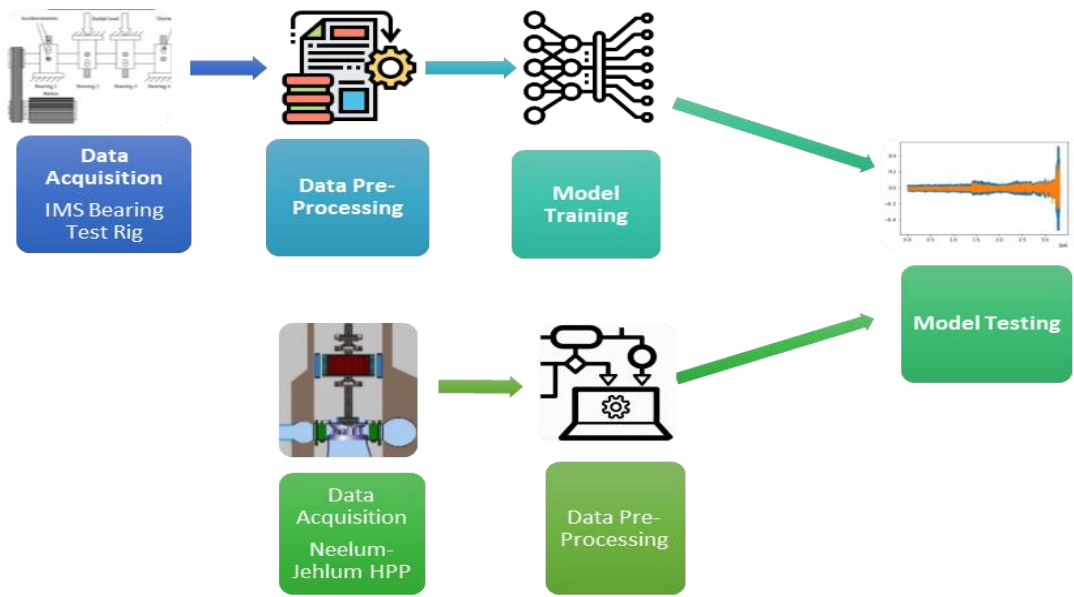

Figure **1** Methodology of the Model

capturing extended temporal dependencies within time series data. This drawback is mitigated by LSTM networks, which employ Forget gates to govern the flow of information between different cell states. This mechanism effectively addresses the processing of lengthy sequences within the data [17]. Consequently, to effectively capture expressive representations and nonlinear dynamic features within time series data, LSTM networks excel over traditional RNNs. This superiority arises from their ability to overcome challenges like vanishing or exploding gradient issues, thus enabling the capture of prolonged data dependencies [2,3]. Considering these advantages, this research is dedicated to constructing a fault prognostic system for bearings. This system integrates LSTM as a machine learning algorithm within the model's framework.

## 2.2. Long-Short Term Memory (LSTM)

The LSTM represents a type of second order RNN network structure acknowledged for its capability to store sequential short-term memories and effectively recall them even after several time-steps [16, 31]. LSTMs incorporate recurrent connections, thereby utilizing the context derived from previous time steps' neuron activations to generate an output. Comprising four essential components, LSTMs manage information flow. The memory cell, or cell state, is responsible for data retention. The forget gate determines the data to be retained or discarded via a sigmoid function. The input gate facilitates the addition of new information or memory cell updates, while the output gate extracts data from the memory cell. Through a tanh function, the information is processed, producing meaningful context that serves as both an output and an input for the subsequent cell. These gate mechanisms operate across the temporal axis, capturing intricate long-term dependencies at each time step. Refer to Figure 2 for an illustration of the fundamental LSTM unit architecture.

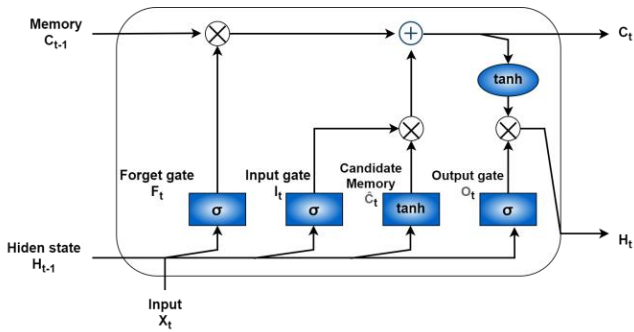

Figure **2** Basic architecture of an LSTM unit

During each time step *t*, the hidden state $H_t$ undergoes an update by combining information from various sources, including the data at the same step $X_t$, the input gate $I_t$, the forget gate $F_t$, the output gate $O_t$, the memory cell $C_t$, and the hidden state from the previous time step $H_{t-1}$. This process is described by the following equations:

$$I_t = \sigma( W_i X_t + V_i h_{t-1} + b_i ) \quad (1)$$

$$F_t = \sigma( W_f X_t + V_f H_{t-1} + b_f ) \quad (2)$$

$$O_t = \sigma( W_o X_t + V_o H_{t-1} + b_o ) \quad (3)$$

$$C_t = F_t \odot C_{t-1} + I_t \odot \tanh( W_c X_t + V_c H_{t-1} + b_c) \quad (4)$$

$$H_t = O_t \odot \tanh( C_t) \quad (5)$$

In equations (1 to 5), the model parameters denoted as $W \in R^{d \times k}$, $V \in R^{d \times d}$, and $b \in R^d$, are learned during training process and are consistently applied across each time step. Where, the sigmoid activation function is represented by **σ**, the element-wise product is denoted by ⊙, and the dimensionality of the hidden layers is defined by a hyper parameter **k.**

The predicted output, which is the future bearing vibration values, is generated through a linear regression layer, which is formulated by the following equation:

$$Y_i = W^r h^T_i \quad (6)$$

In Equation 6, the model predicts the sixth value using the first five values (1-5) from the input, and subsequently predicts the seventh value using input values (2-6), and so on.

The dimensionality of the output are represented by $W^r \in R^{k \times z}$. $W^r$ is the weight matrix associated with the reset gate, having **k** rows and **z** columns**.** During the model training, cross-entropy serves as the loss function, measuring the disparity between the desired target label distribution **p(x)** and the predicted label distribution **q(x)**. The cross-entropy between **p(x)** and **q(x)** is given by:

$$\text{Loss} = H(p,q) = -\sum_x p(x)\log(q(x)) \quad (7)$$

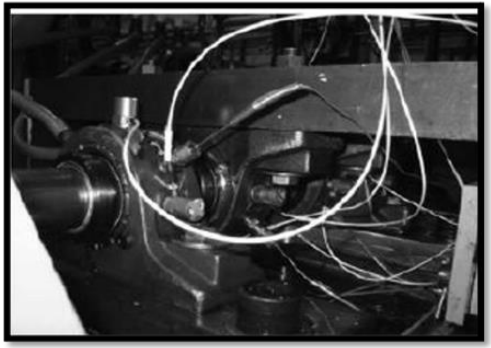
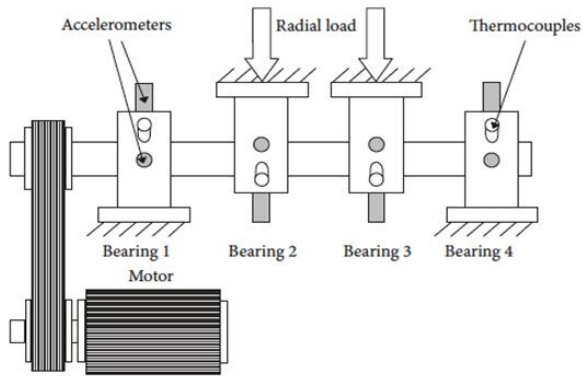

Figure 3 Test Rig by IMS [32]

The activation function plays a crucial role in allowing the network to capture nonlinear patterns within the input signal. This ability enhances the network's capacity to learn discriminative features that contribute to its overall representation power. LSTM, on the other hand, capitalizes on both the spatial and temporal attributes present in raw temporal data, mirroring the memory mechanisms of the human brain. This unique characteristic, positions LSTM-based architectures to potentially achieve greater accuracy in the realm of fault prognostics.

## 2.3. Data Acquisition and Pre-processing

During this research, two distinct sets of data have been acquired. The initial set encompasses bearing vibration data obtained from a test rig, while the subsequent set encompasses authentic data sourced from an operational hydro power project located in Pakistan. Additional details regarding these datasets are expounded upon in the subsequent sections.

### 2.3.1. IMS Dataset

In the initial phase, the model was trained employing the dataset furnished by the Intelligent Maintenance System (IMS) at the University of Cincinnati [32]. The comprehensive particulars of the IMS Dataset test rig setup are illustrated in Table 1.

Table **1** Detailed description of IMS Dataset Test Rig

| | |
|---|---|
| **Bearings Type** | **Double Rows Rexnord ZA-2115** |
| **Total Bearings** | 04 |
| **Total Load on Shaft** | 6000 lbs |
| **Rotational Speed of the Shaft** | 2000 rpm |
| **Accelerometers Type** | High Sensitivity Quartz ICP |
| **Total Accelerometers (Test 01)** | 02 Accelerometers on x and y axis |
| **Total Accelerometers (Test 02 & Test 03)** | 01 Accelerometer |
| **Sampling Rate** | 20 kHz |

The configuration of the test rig's layout is presented in Figure 3. A system for oil circulation was implemented to ensure lubrication within the test rig. The oil feedback pipe was equipped with a magnetic switch designed to capture debris, aiding in the identification of bearing degradation. Data recording took place at 10-minute intervals, each span lasting 1 second. Every recorded data file contained 20480 data points, and the specific recording time was indicated in the filename. This experiment followed a run-to-failure approach, with data collection persisting until a fault was introduced in the bearings. The dataset 2 and dataset 3, at the end of the run-to-failure experiments recorded an outer race fault in the Bearing 1 and Bearing 3, as illustrated in Figure 4.

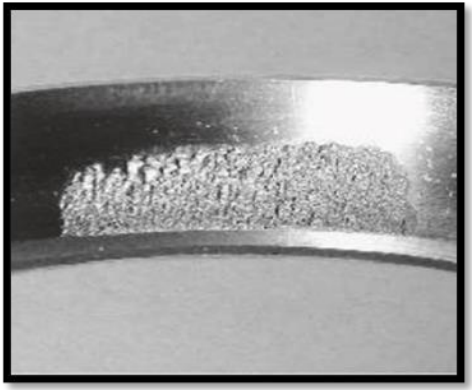

Figure **4** Outer race fault in the bearings [33]

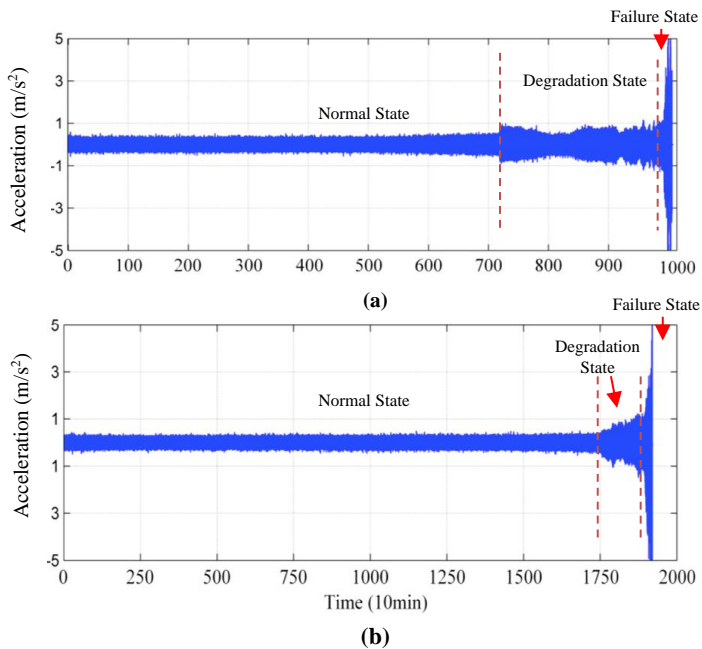

Figure **5** Plot of bearing vibrational signals (IMS)

Figure 5 (a) and Figure 5 (b) depict the plot of vibration signals for bearing 1 and bearing 3 as recorded for the entire run-to-failure experiment in dataset 2 and dataset 3, respectively. It is clear from the plot that the bearings were first operating in a normal state until a slight degradation in their state where the vibration values have been increased. Followed by a failure state where physical wear and tear was recorded in the bearing and a further increase in the vibration values as evident from the plot.

### 2.3.2. Neelum-Jehlum Hydropower Project Dataset

Bearing vibration data for a period of twelve months, recorded by the SCADA system installed at 969MW NJHPP Pakistan, have been acquired. Four number of units (turbines), each having a generation capacity of 242.25 MW are installed at NJHPP [30]. The data used in this research was recorded from the horizontal vibration runout system of the Turbine Guide Bearing installed at Unit No. 01. The reason being a fault had occurred in the Turbine Guide Bearings of Unit 01 during operation, hence the data contained both clean and faulty data points.

Furthermore, the reason for acquiring twelve-month data is to encompass both the lean water period where turbines face a lot of turbulence and the peak season where turbines normally operate seamlessly. Figure 6 shows the plot of the overall recorded data.

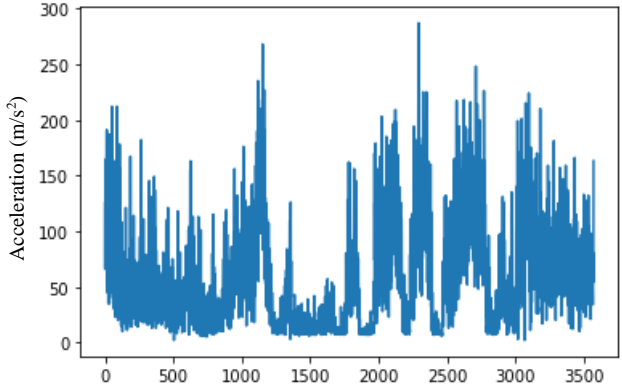

Figure 6 Plot of bearing vibrational signal (NJHPP)

### 2.3.3. Data Preprocessing

The pre-processing of the data to mitigate concerns related to noise, data redundancy, and missing values. Following this, an outlier removal procedure was implemented. Due to the differing ranges of the data collected from both sources, normalization was performed using the Minmax-scaler function outlined in Equation 8.

$$x' = \frac{x - \min(x)}{\max(x) - \min(x)} \quad (8)$$

Where the original value is represented by x and the normalized value is represented by x'. Rescaling the data had a dual impact – not only did it contribute to improving the model's ability to generalize, but it also expedited the learning process and facilitated faster convergence rates.

### 2.4. Model Training and Testing

The IMS dataset consists of three distinct run-to-failure experiments. Following experiment 02 and experiment 03, based on dataset 2 and 3, consecutively, outer-race faults were experienced in bearing 1 and bearing 3, respectively. Therefore, dataset 02 serves the dual purpose of training and testing the model. Dataset 03, however, is only used for testing to provide an additional layer of performance validation.

Table 2 Parameters of the LSTM Model

| Model Type | Stacked LSTM |
|---|---|
| Total Hidden Layers | 02 |
| Total Memory Units | *Layer One:* 128 *Layer Two:* 64 |
| Optimizer | Adam |
| Learning Rate | 0.001 |
| Batch Size | 50 |
| No. of Epochs | 100 |

To begin, the 984 files within dataset 02 are divided into training and testing sets, maintaining a 70:30 ratio. This allocation dedicates 70% data to model training and 30% to model testing. The model's performance is subsequently assessed using dataset 03, which comprises 4448 files. Once the model has undergone training, testing, and fine-tuning with the IMS dataset, its evaluation extends to bearing vibration data from NJHPP. Table 2 showcases the parameters of the developed stacked LSTM model.

The evaluation of the proposed model's performance relies on the Root Mean Square Error (RMSE), calculated using equation 9. This choice of metric is driven by the consideration that significant errors can lead to unfavorable outcomes in prognostics. Hence, RMSE proves to be a valuable evaluation measure due to its emphasis on larger error values.

$$RMSE = \sqrt{\frac{1}{n}\sum_{i=1}^{n}(y_i - \hat{y}_i)^2} \quad (9)$$

Where, $y_i$ represents the actual values and $\hat{y}_i$ represents the predicted values.

### 3. RESULTS AND DISCUSSION

The model's performance was evaluated using both the IMS dataset and the Neelum Jhelum dataset. The outcomes demonstrate that the model effectively forecasted future bearing vibration values, yielding remarkably low RMSE scores. The testing conducted on data from two distinct sources also affirms the model's ability to generalize well. Regardless of the data's origin, the model exhibited efficient predictive capabilities for bearing vibrations, thereby mitigating the necessity for extensive domain expertise.

## 3.1. Test Results – IMS Dataset

The model's performance was evaluated against dataset 02 and dataset 03. The graphical representation of predicted and actual vibration values is illustrated in Figure 7 (a) and (b), where blue indicates actual values and orange denotes predicted values. The plot clearly demonstrates the model's accurate prediction of bearing vibrations, tracking the degradation trend until the fault occurrence. Additionally, the RMSE values for dataset 02 and dataset 03 were calculated as 0.0145 and 0.0102, respectively.

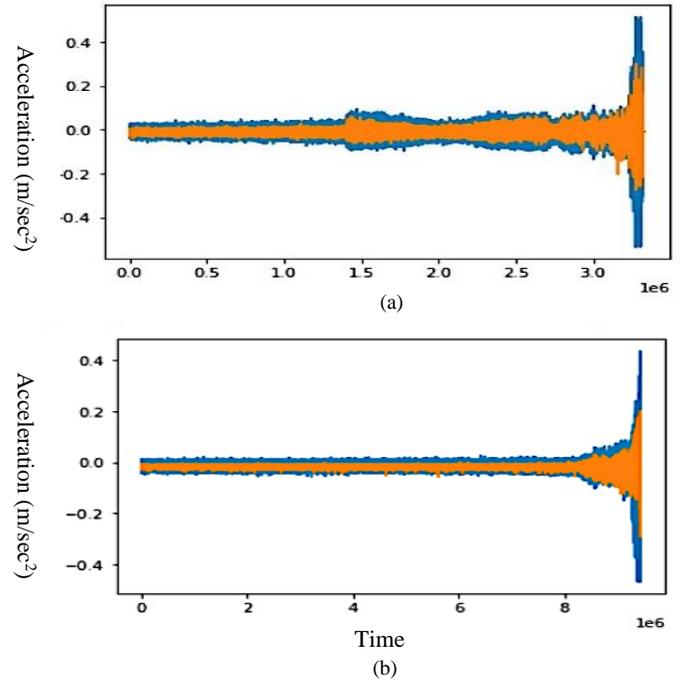

Figure **7** Bearing Vibrational Signals – Actual vs Predicted Values (IMS Dataset) [6]

Table **3** Model Evaluation Using Various Metrices

| Model | RMSE | MAE | NMAE | MAPE |
|---|---|---|---|---|
| LSTM using raw bearing vibration values | 0.0102 | 0.0108 | 0.0002 | 0.0107 |

These RMSE values signify a minimal disparity between actual and predicted bearing vibration values, validating the precision of the developed model. Furthermore, during the testing on dataset 03, the performance of the model was evaluated using other metrics including Mean Absolute Error (MAE), Normalized Mean Absolute Error (NMAE) and Mean Absolute Percentage Error (MAPE). The results are depicted in Table 3

## 3.2. Test Results – NJHPP Dataset

The model's performance was evaluated using data from the NJHPP, and the outcomes are illustrated in Figure 8. This figure displays the graph of the actual and predicted vibration values of the bearings. The actual vibration values is indicated by blue color, whereas the green and red colors correspond to predictions during the training and testing phases, respectively.

The plot clearly demonstrates that the model accurately predicts both normal and faulty bearing vibrations, closely following the observed trends. Furthermore, the RMSE of the predicted values for the NJHPP dataset was exceptionally low, measuring only 0.11. This result indicates the model's effectiveness in accurately forecasting vibration data values, regardless of the data source.

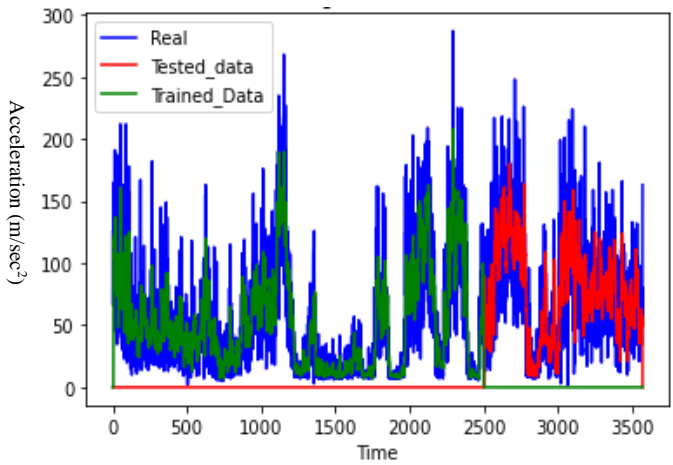

Figure **8** Bearing Vibrational Signals–Actual vs Training and Testing Predicted Values (NJHPP Dataset) [6]

The analysis underscores the promising outcomes of the developed LSTM model. The remarkable accuracy in predicting vibration data not only underscores its efficacy but also diminishes the necessity for profound domain expertise. This discovery indicates that the model possesses significant capabilities and adaptability in handling diverse data sources, rendering it a valuable asset across various applications.

## 4. CONCLUSION

This research primarily focuses on the development of an LSTM-based model for conducting fault prognostics in hydropower plant bearings. LSTM holds an advantage over other algorithms due to its capacity to diminish the necessity for expert domain knowledge and intricate feature engineering. This advantage arises from its deep architecture and hierarchical feature extraction, which endows the learning model with the potential to effectively discern intrinsic patterns within time series data. Moreover, the incorporation of forget gates in LSTM enables the capture of long-term dependencies. Consequently, LSTM adeptly identifies and unveils significant features within sensory signals while executing bearing fault prognostics in hydropower plants (HPPs).

The proposed model is meticulously trained and tested using the IMS dataset derived from a dedicated test rig. The outcomes of these tests showcase that the model has achieved remarkably low RMSE values and proficiently projected the forthcoming vibration states of the bearings, accurately detecting faults as well. In the subsequent research phase, the model's performance is evaluated using real vibration data collected from the SCADA system of the Neelum Jhelum Hydropower Plant (NJHPP) operating in Pakistan. Once again, the results underscore that the model has attained a remarkably low RMSE value and effectively anticipated and traced the trends within bearing vibration data.

Given that the developed LSTM model consistently attains impressive outcomes by precisely forecasting future bearing vibration conditions within a hydropower plant, it holds the potential to empower HPP operators to anticipate such faults in advance. This not only curtails maintenance expenses but also ensures continuous plant availability—a crucial aspect, particularly in developing nations like Pakistan. In such regions, where electricity demand outpaces supply and unscheduled plant downtimes further jeopardize energy security, this research's findings carry substantial significance. Additionally, while this research's scope pertains primarily to hydropower projects, its applicability can extend to other green energy generation ventures, such as wind power projects, considering that bearings also play a pivotal role in wind turbines.

## Statements and Declarations

In the interest of full transparency, we disclose that the authors have no competing interests related to this research. There are no financial, personal, or professional conflicts of interest that could influence the interpretation or presentation of the findings in this manuscript.

## Data Availability Statement:

The data used to support the findings of this study have been deposited in the figshare repository [35]. (https://doi.org/10.6084/m9.figshare.21290895).

## REFERENCES

1. Hannah Ritchie and Pablo Rosado (2020) - "Energy Mix" Published online at OurWorldInData.org. Retrieved from: 'https://ourworldindata.org/energy-mix' [Online Resource]

2. A Review on Data-driven Predictive Maintenance Approach for Hydro Turbines/Generators (International Workshop of Advanced Manufacturing and Automation (IWAMA 2016))

3. Yu, L., Qu, J., Gao, F. and Tian, Y., 2019. A novel hierarchical algorithm for bearing fault diagnosis based on stacked LSTM. Shock and Vibration, 2019.

4. Markova, M., 2022, April. Convolutional neural networks for forex time series forecasting. In AIP Conference Proceedings (Vol. 2459, No. 1). AIP Publishing

5. Ge, Y., Guo, L. and Dou, Y., 2019. Remaining useful life prediction of machinery based on KS distance and LSTM neural network. *International Journal of Performability Engineering*, *15*(3), p.895.

6. Afridi, Y.S.; Hasan, L.; Ullah, R.; Ahmad, Z.; Kim, J.-M. LSTM-Based Condition Monitoring and Fault Prognostics of Rolling Element Bearings Using Raw Vibrational Data. Machines 2023, 11, 531.

7. H. Jiang, C. Li, and H. Li, "An improved EEMD with multiwavelet packet for rotating machinery multi-fault diagnosis," Mechanical Systems and Signal Processing, vol. 36, no. 2, pp. 225–239, 2013.

8. Y. Li, X. Liang, and M. J. Zuo, "Diagonal slice spectrum assisted optimal scale morphological lter for rolling element bearing fault diagnosis," Mechanical Systems and Signal Processing, vol. 85, pp. 146–161, 2017A Review on Prognosis of Rolling Element Bearings (N.S. Jammu et al. / International Journal of Engineering Science and Technology (IJEST)

9. Susto, G.A., Schirru, A., Pampuri, S., McLoone, S. and Beghi, A., 2014. Machine learning for predictive maintenance: A multiple classifier approach. *IEEE Transactions on Industrial Informatics*, *11*(3), pp.812-820.

10. Y. Goldberg, "A primer on neural network models for natural language processing," Computer Science, 2015, http://arxiv.org/abs/1510.00726.

11. C. Szegedy, V. Vanhoucke, S. Ioffe, J. Shlens, and Z. Wojna, "Rethinking the inception architecture for computer vision," in Proceedings of IEEE Conference on Computer Vision and Pattern Recognition, Las Vegas, NV, USA, June 2016.

12. O. Janssens, V. Slavkovikj, B. Vervisch et al., "Convolutional neural network based fault detection for rotating machinery," Journal of Sound and Vibration, vol. 377, pp. 331–345, 2016.

13. Z. Q. Chen, C. Li, and R. V. Sanchez, "Gearbox fault identification and

classification with convolutional neural networks," Shock and Vibration, vol. 2015, Article ID 390134, 10 pages, 2015.
14. L. Guo, H. L. Gao, Y. W. Zhang et al., "Research on bearing condition monitoring based on deep learning," Journal of Vibration and Shock, vol. 35, no. 12, pp. 166–171, 2016
15. Z. Song, K. Wu, J. Shao **Destination prediction using deep echo state network**Neurocomputing., 406 (2020),pp. 343353, 10.1016/j.neucom.2019.09.115.
16. S. Hochreiter, J. Schmidhuber Long short-term memory Neural Comput., 9 (1997), pp. 1735-1780
17. Mikolov, T., Joulin, A., Chopra, S., Mathieu, M. and Ranzato, M.A., 2014. Learning longer memory in recurrent neural networks. *arXiv preprint arXiv:1412.7753*.
18. R. Girshick, J. Donahue, T. Darrell and J. Malik, "Region-based convolutional networks for accurate object detection and semantic segmentation", *IEEE Trans. Pattern Anal. Mach. Intell.*, vol. 38, no. 1, pp. 142-158, Jan. 2016
19. K. He, X. Zhang, S. Ren and J. Sun, "Spatial pyramid pooling in deep convolutional networks for visual recognition", *IEEE Trans. Pattern Anal. Mach. Intell.*, vol. 37, no. 9, pp. 1904-1916, Sep. 2015.
20. Z. He, J. Chen, Y. Zi, and J. Pan, "Independence-oriented VMD to identify fault feature for wheel set bearing fault diagnosis of high speed locomotive," Mechanical Systems and Signal Processing, vol. 85, pp. 512–529, 2017.
21. I. M. Jamadar and D. P. Vakharia, "A novel approach integrating dimensional analysis and neural networks for the detection of localized faults in roller bearings," Measurement, vol. 94, pp. 177–185, 2016.
22. X. Xia, J. Zhou, J. Xiao, and H. Xiao, "A novel identification method of Volterra series in rotor-bearing system for fault diagnosis," Mechanical Systems and Signal Processing, vol. 66- 67, pp. 557–567, 2016.
23. Feng, B., Zhang, D., Si, Y., Tian, X. and Qian, P., 2019, September. A condition monitoring method of wind turbines based on Long Short-Term Memory neural network. In *2019 25th International Conference on Automation and Computing (ICAC)* (pp. 1-4). IEEE.
24. Zhao, R., Wang, J., Yan, R. and Mao, K., 2016, November. Machine health monitoring with LSTM networks. In *2016 10th international conference on sensing technology (ICST)* (pp. 1-6). IEEE.
25. Wang, J., Wen, G., Yang, S. and Liu, Y., 2018, October. Remaining useful life estimation in prognostics using deep bidirectional LSTM neural network. In *2018 Prognostics and system health management conference (PHM-Chongqing)* (pp. 1037-1042). IEEE.
26. Huang, C.G., Huang, H.Z. and Li, Y.F., 2019. A bidirectional LSTM prognostics method under multiple operational conditions. *IEEE Transactions on Industrial Electronics*, *66*(11), pp.8792-8802..
27. Zhao, R., Yan, R., Wang, J. and Mao, K., 2017. Learning to monitor machine health with convolutional bi-directional LSTM networks. *Sensors*, *17*(2), p.273
28. Le, X.H., Ho, H.V., Lee, G. and Jung, S., 2019. Application of long short-term memory (LSTM) neural network for flood forecasting. *Water*, *11*(7), p.1387.

29. Yuhang, C.H.E.N. and Bing, H.A.N., 2019, December. Prediction of bearing degradation trend based on LSTM. In *2019 IEEE Symposium Series on Computational Intelligence (SSCI)* (pp. 1035-1040). IEEE.
30. WAPDA, 2012. Neelum Jhelum Hydropower Project(http://wapda.gov.pk/vision2025/htmls_ with-energy-crisis-in-pakistan/). vision2025/njhp.html).
31. Monner, D. and Reggia, J.A., 2012. A generalized LSTM-like training algorithm for second-order recurrent neural networks. *Neural Networks*, *25*, pp.70-83.
32. J. Lee, H. Qiu, G. Yu, and J. Lin, "Rexnord Technical Services,"Bearing Data Set", IMS, University of Cincinnati. NASA Ames Prognostics Data Repository," NASA Ames, Moffett Field, CA, https://ti.arc.nasa.gov/tech/dash/groups/pcoe/prognostic-data-repository/.
33. Wavelet filter-based weak signature detection method and its application on rolling element bearing prognostics, Qiu, Hai and Lee, Jay and Lin, Jing and Yu, Gang, Journal of Sound and Vibration, Vol. 289 No. 4, 1066--1090, 2006
34. Afridi YS, Ahmad K, Hassan L. Artificial intelligence based prognostic maintenance of renewable energy systems: A review of techniques, challenges, and future research directions. Int J Energy Res. 2021;1-24.
35. Afridi, Yasir Saleem (2022): Bearing Vibration Dataset of a Hydropower Project. figshare. Dataset. https://doi.org/10.6084/m9.figshare.2129089.
36. Afridi, Y.S., Hassan, L., Ahmad, K. (2023). Machine Learning Applications for Renewable Energy Systems. In: Manshahia, M.S., Kharchenko, V., Weber, GW., Vasant, P. (eds) Advances in Artificial Intelligence for Renewable Energy Systems and Energy Autonomy. EAI/Springer Innovations in Communication and Computing. Springer, Cham. https://doi.org/10.1007/978-3-031-26496-2_5
37. Wen, Q., Zhou, T., Zhang, C., Chen, W., Ma, Z., Yan, J. and Sun, L., 2022. Transformers in time series: A survey. arXiv preprint arXiv:2202.07125.
38. Zhou, H., Zhang, S., Peng, J., Zhang, S., Li, J., Xiong, H. and Zhang, W., 2021, May. Informer: Beyond efficient transformer for long sequence time-series forecasting. In Proceedings of the AAAI conference on artificial intelligence (Vol. 35, No. 12, pp. 11106-11115).
39. Kim, J., Oh, S., Kim, H. and Choi, W., 2023. Tutorial on time series prediction using 1D-CNN and BiLSTM: A case example of peak electricity demand and system marginal price prediction. Engineering Applications of Artificial Intelligence, 126, p.106817.